# Parallel Deep Learning-Driven Sarcasm Detection from Pop Culture Text and English Humor Literature

Sourav Das[1][0000-0003-4864-1635] and Anup Kumar Kolya[2][0000-0002-8758-7670]

[1] National Institute of Electronics & Information Technology, Kolkata 700032, India
[2] Dept. of Computer Science, RCC Institute of Information Technology, Kolkata 700015, India
{sourav.das.research,anup.kolya}@gmail.com

**Abstract.** Sarcasm is a sophisticated way of wrapping any immanent truth, message, or even mockery within a hilarious manner. The advent of communications using social networks has mass-produced new avenues of socialization. It can be further said that humor, irony, sarcasm, and wit are the four chariots of being socially funny in the modern days. In this paper, we manually extract the sarcastic word distribution features of a benchmark pop culture sarcasm corpus, containing sarcastic dialogues and monologues. We generate input sequences formed of the weighted vectors from such words. We further propose an amalgamation of four parallel deep long-short term networks (pLSTM), each with distinctive activation classifier. These modules are primarily aimed at successfully detecting sarcasm from the text corpus. Our proposed model for detecting sarcasm peaks a training accuracy of 98.95% when trained with the discussed dataset. Consecutively it obtains a highest of 98.31% overall validation accuracy on two handpicked Project Gutenberg English humor literature among all the test cases. Our approach transcends previous state-of-the-art works on several sarcasm corpora and results in a new gold standard performance for sarcasm detection.

## 1 Introduction

Sarcasm used in any language relies heavily on the context of the subjectivity being discussed. Thus, it can be delivered without showing the signs of any expressions whatsoever, as keeping a straight face, a smirk, or even with laughter. It is a challenge to extract the actual underlying meaning of a sarcastic statement, rather than finding the straight targeted sentiment polarity. However, when an independent sarcastic text corpus is cultured, we may not have to consider the whole surroundings and context, but with the sarcastic texts only.

Most of the research work in this area is conducted with binary classification tasks and empirical details to detect any text or phrase that is sarcastic or not [1]. Also, the accuracy of neural networks deployed for sarcasm detection tasks is dependent on the context or sentence level attention covering an entire sarcasm corpus [2, 3]. Presently deep learning networks are represented as the computational combinations of weighted vectors through input neurons. Hence the training can improve their ability to reproduce the weightage combinations over time, enhancing the overall learnability and detection capability of sarcasm, but often fails to accumulate the subjectivity. In this work, we aim to accurately identify the complex sarcasm within spoken dialogues translated to texts. At first, we take up a sarcasm text corpus for reading and manually tokenizing the file for lexical occurrences indexing within the file. Secondly, we channelize the weighted vectors gained from the dataset and combine our LSTM networks by feeding the input with such vectors from the token indices. We introduce a simple yet robust set of four parallel long short-term networks (pLSTM) with dense hidden layers, while each set consisting of a different activation function. Finally, the proposed network is evaluated against a handpicked number of open-source humor and sarcasm text corpora. For all the generic validation cases, our cross-activation classifier incorporated parallel long-short term networks to achieve a better test accuracy of 98.31% when compared with other diverse data and models for a similar task. The rest of the sections are organized as follows: In Section 2, some recent works are mentioned in a similar domain. In Section 3, we discuss the attributes of the dataset used, with the feature extraction for sarcastic words. In Section 4, we introduce and formulate our proposed parallel long-short term networks. We state results and analysis regarding validation and classification in Section 5. In Section 6, the performance of our approach is compared against several standard humor and sarcasm datasets. Section 7 consists of a discussion on a few pivotal points from our work. Finally, we consider the future extensions of the work and conclude in Section 8.

## 2 Recent Works

Recent trends in deep neural networks have opened up diversified applications in sarcasm detection. In accordance, a group of researchers proposed sentiment classification and sarcasm detection as the correlated parts of a common linguistic challenge [4]. For their said approach, they chalked out multiple task-based learning systems, respectively the sentiment and sarcasm classification tags. They combined the gated network with a fully connected layer and softmax activation and observed that their proposed classifier utilizes the sentiment shift to detect sarcasm better. Kumar et al. exploited attention-based bidirectional and convolutional networks for sarcasm detection from the benchmark dataset [5]. They selected an already developed sarcastic tweets dataset and randomly streamed tweets on sarcasm. They introduced an attention layer constructed within an LSTM network, embedding the softmax activation function embedded into the backpropagation property. The backpropagation-based feedback helps to identify and differentiate tweets from each other. Sundararajan and Palanisamy classified sarcasm detection in different genres, naming polite, rude, raging types of sarcasm [6]. Instead of defining the straight polarity of a sarcastic expression, they extracted the mixed emotions associated with the sarcasm itself. They fragmented several feature parameters of the input tweets and ensembled them for a better semantic understanding of the tweets. Finally, they applied a rule-based classifier to eliminate the fuzziness of the inputs.

## 3 Corpus

To train our model, we select the text corpus from the MUStARD or Multimodal Sarcasm Detection Dataset [7]. As the name suggests itself, the dataset is comprised of multimodal aspects of sarcastic expressions, i.e., videos, audios, and text corpus accumulated from the dialogues. The data contains a collection of 6 thousand videos from several popular TV comedy shows. Following that, the utterance of videos was manually annotated, defining in which context the characters delivered sarcastic dialogues. Furthermore, from the entire range of their video repository, they selected only above 600 videos as a blend of balanced sarcastic and non-sarcastic labeled. From the multi-modality parameters of dialogues' utterance, only the transcription or textual modality is selected. The researchers' overall character-label ratio and distribution help us to visualize which characters have the most utterance of words throughout the dialogues gathered. Also, what part of dialogue contributions the characters have in building the corpus. The final proposed textual corpus consists of 690 lines of individual line indexed dialogues. The highest dialogue per character cumulation reaches a maximum of up to uninterrupted 65 words, whereas the minimum dialogue ranging is 7 words. It symbolizes that the corpus has well consisted of both monologues and dialogues. To represent the contextual text utterance, the researchers represented sentence level utterance features from BERT [8]. Finally, the dataset showcases the occurrences of sarcastic word utterances belong to the respective sentences, alongside the first token average for the sentences.

### 3.1 Sarcastic Words Distribution

As the entire corpus highly consists of sarcastic statements, we do not partition balanced or imbalanced data allocations further. We primarily focus to extract the frequency of the hundred most recurring words in dialogues. Since it is not a plain documented text from literature, rather consists of sarcastic dialogues, it can be inferred that each time these words occur within a dialogue or monologue, is intended for saying something sarcastic. Such few most occurred words are shown from the entire corpus in Table 1.

**Table 1.** A few of the most frequent words with their frequency span within the corpus.

| Words | Frequency | Distribution (%) |
|:---:|:---:|:---:|
| 'oh' | 74 | 8.21 |
| 'know' | 49 | 7.90 |
| 'like' | 46 | 7.78 |
| 'yeah' | 43 | 7.66 |
| 'well' | 39 | 7.52 |
| 'go' | 37 | 7.50 |
| 'right' | 34 | 7.47 |
| 'think' | 32 | 7.46 |
| 'really' | 30 | 7.44 |

## 4  Deep pLSTM Architecture

We propose to construct four equipollent long-short term networks for parallel learning with identical sizes as our combined baseline model. We term them as pLSTM networks, with a fixed batch of inputs but individual outputs. Each LSTM network consists of distinguished end classifier functions at the output. These classifiers, or generically termed as activation functions, help us to understand the analogous comparison of their behavioral natures while handing large input vectors generated from fragmenting the text corpus. We channelize the input vectors generated from tokens from the text data in a four-parallel way as the input feeding for each corresponding model. Each of these models contain fully connected deep layers, but isolated from each other within the architecture.

The proposed networks consist of the bidirectional signaling within their layers, containing a backward LSTM pass and a forward LSTM pass, for the forget to get back-propagation, as well as the output gate serving as the input for the next layer. The combination of a forward and backward pass within an LSTM layer can be standardized formally as:

For incoming input into a layer:

$$i = \Phi \left( W_i x_i + U_i (F_{seq(\rightarrow)} + R_{seq(\leftarrow)}) + b_i \right) \quad (1)$$

Where $\Phi$ denotes the activation function of the respective network, $W$ is the weight vector of input in the layer, $U$ can be denoted as the updated combination of sequence signals, and $b$ is the bias vector for output summarization of that particular layer.

Similarly, for the output generation from a layer:

$$o = \Phi \left( W_o x_o + U_o (F_{seq(\rightarrow)} + R_{seq(\leftarrow)}) + b_o \right) \quad (2)$$

For the forget gate of a layer:

$$f = \Phi \left( W_f x_f + U_f (F_{seq(\rightarrow)} + R_{seq(\leftarrow)}) + b_f \right) \quad (3)$$

Finally, the computation cell first takes the output quality of the forget gate $f$ and simultaneously sustain the previous layer's input characteristics, or forget them. Similarly, it takes the input coming from the of the input gate $i$, and accordingly channels it as the new computational memory as $\tilde{c}$. It then sums these two results to produce the final computation memory $c$.

Hence, for the computation cell operation:

$$c = (f (W_n + U_n) \cdot c) + (i_n \cdot \tilde{c}) \quad (4)$$

Now, these are the representative modelling of a respective layer of a single long-short term network. Obviously, as the layering densifies, these equational simulations will be clubbed together for each layer. Henceforth, we sum up together the inputs from eq. (1) for each LSTM channel as:

$$i_{1...n} = [\Phi \text{ (softmax)} (W_i x_i + U_i (F_{seq(\rightarrow)} + R_{seq(\leftarrow)}) + b_i)] +$$
$$i_{2...n} = [\Phi \text{ (sigmoid)} (W_i x_i + U_i (F_{seq(\rightarrow)} + R_{seq(\leftarrow)}) + b_i)] +$$
$$i_{3...n} = [\Phi \text{ (relu)} (W_i x_i + U_i (F_{seq(\rightarrow)} + R_{seq(\leftarrow)}) + b_i)] +$$
$$i_{4...n} = [\Phi \text{ (tanh)} (W_i x_i + U_i (F_{seq(\rightarrow)} + R_{seq(\leftarrow)}) + b_i)] \quad (9)$$

Here, *n* is the notation for the number of inherent layers within each respective LSTM structure. Similarly, for the output cumulation of fragments within the unified architecture $o_1 \rightarrow o_4$, forget gates $f_1 \rightarrow f_4$, and $c_1 \rightarrow c_4$ are summarized for each layer.

### 4.1 Hyperparameters Tuning

The model tuning contains a word embedding dimension of 400. We further deploy 500 hidden layers for each of the deep LSTM substructures within the composite architecture. The dense layer holds different activation classifiers for each substructure. The *adam* optimizer is used with categorical cross-entropy for binary categorical classification of sarcasm within the training phase. It is then initialized with the loss function to evaluate the training loss. The learning rate of the optimizer is kept as 0.01. The dropout rate for avoiding overfitting is respectively 0.6 for the vectorization and 0.4 for the bidirectional layers. The epochs for each LSTM module are set as 500. The verbose information is kept as 1 for words (vectors) to training logs. Finally, the collective outcome is printed as *model.summary( )*.

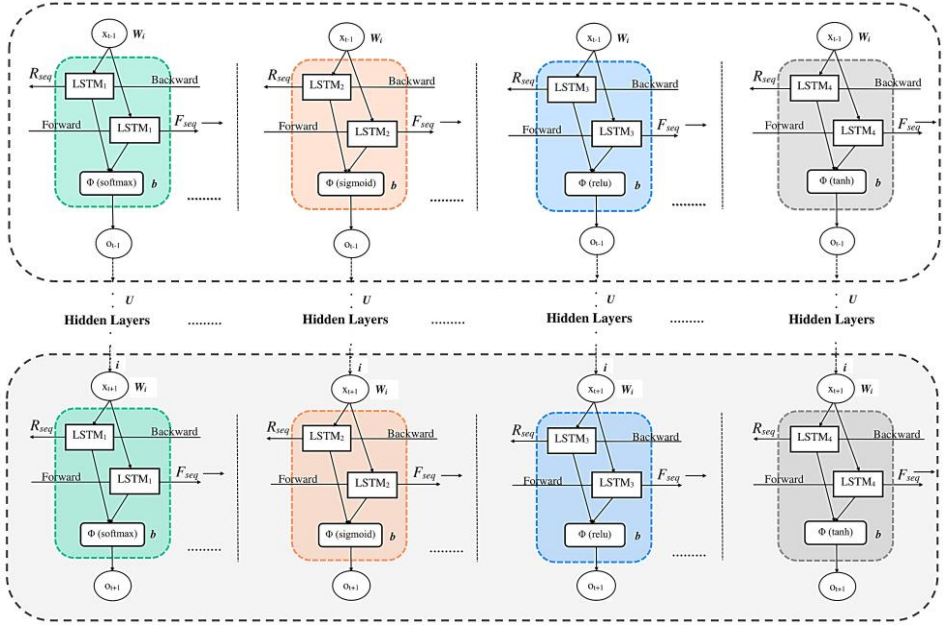

**Fig. 1.** Proposed architecture for parallel long-short term networks.

## 5  Results and Analysis

At first, we compare the performance of four pLSTM modules with distinct activation classifiers. The comparison is made while all the modules reach the peak training locus assigned, i.e., on the 500[th] epoch. We show the scalable comparison graph in Fig. 2.

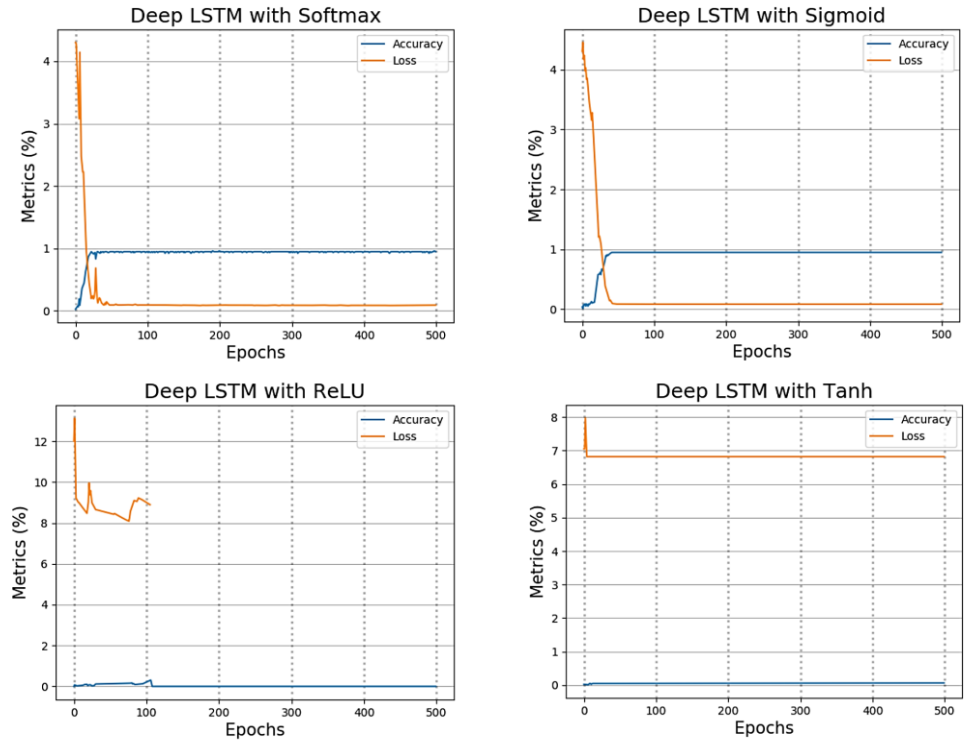

**Fig. 2.** Epoch level analysis of pLSTM modules with training loss and accuracy.

From Fig. 2, it can be observed that our baseline pLSTM module with softmax activation function minimizes down the training error to a minimal range even before the first 100 epochs. The training accuracy also peaks above 90% during this phase. Hence an overlap takes place between 0 to 100 epochs, where the error rate comes down while the accuracy emerges. Meanwhile, the accuracy simultaneously goes up to almost touching the peak margin of 98.95%, and it mostly settles within the high accuracy range of 96%-98% for the rest of the training phase. The LSTM coupled with sigmoid performs similarly, but closer analytical observation reveals it tops an accuracy of 96.88%, narrowly falling short of the first module. On the other hand, LSTM modules coupled with ReLU and tanh both suffer from a heavy range of data loss during training and perform drastically worse. Since the loss values remain constantly high, the training success for base sarcasm classification does not progress much. Table 2 represents the training analysis concerning epochs. Followingly, we discard the train sets of both LSTM coupled with ReLU and tanh. We further compare the first two classifier combined modules in Table 3, where performance visibility helps to narrow down the F-measure analysis. Here the performance represents the classification report along with the F-score generated from the training phase.

**Table 2.** Summary information for training sets, where *S/NS* is the successful detection of sarcastic or non-sarcastic dialogues and monologues from the data per 100 epochs.

| Modules | S/NS 100 | S/NS 200 | S/NS 300 | S/NS 400 | S/NS 500 |
|---|---|---|---|---|---|
| pLSTM + softmax | 93.07 | 92.15 | 95.33 | 97.79 | **98.95** |
| pLSTM + sigmoid | 91.12 | 90.03 | 92.50 | 94.52 | **96.88** |
| pLSTM + relu | 21.27 | 12.20 | 11.16 | 11.90 | 10.63 |
| pLSTM + tanh | 10.60 | 09.23 | 09.11 | 08.25 | 08.16 |

**Table 3.** Comparative evaluation of the better performing pLSTM modules w.r.t classification report.

| Modules | Sarcasm | | | |
|---|---|---|---|---|
| | Precision | Recall | F1-Score | Accuracy |
| pLSTM + softmax | 0.9900 | 0.9800 | 0.9851 | 0.9850 |
| pLSTM + sigmoid | 0.9600 | 0.9400 | 0.9505 | 0.9500 |

## 6  Benchmark Comparisons

We carry out similar rounds of experimentation with several open-sourced and sarcasm-based datasets for evaluating generic validation accuracy. At first, we select two works of English humor and comedy literature from the Project Gutenberg digital library[1]. These are The Comedy of Errors and Three Men in a Boat. Each of them represents linear forms of plain texts, without the need for any preprocessing. Followingly, we examine our proposed framework with three more sarcasm data scraped and developed from the internet. First of which is a sarcastic and non-sarcastic combination of data curated by a supervised pattern (SIGN) [9]. The main texts in the corpus are questions and rhetorical statements, in quote and response pairs. The second dataset developed is a sarcastic and non-sarcastic combination of data curated by a supervised pattern (Sarcasm V2) [10]. The final corpus selected is a collection of train and test data on sarcasm from Reddit posts (SARC) [11]. It is a manually annotated corpus primarily consisting of eight years of 1.3 million sarcastic Reddit posts, their replies, and comments.

We shuffle the data by performing 5-fold random cross-training from each dataset, in a 3:2 manner. Then again, we apply the testing on the entire corpora. Table 4 shows that among all the sarcasm datasets, our baseline pLSTM with softmax obtains the highest validation accuracy of 98.31%, gaining a lead margin of 2.27% from the next best performing substructure of pLSTM with sigmoid (96.04%). Not only that, but both the softmax and sigmoid attached modules also obtain F scores 97.03 and 94.58 respectively for TCE based validation, which is the highest among the benchmark analysis. For [9, 10], our proposed method outperforms the previous state-of-the-art results [11, 12] by a notable improvement leap. It is evident from the results that the feature vectors fed as input for the parallel LSTMs do not lead to overfitting, and the deep structured module(s) utilizes them to almost the saturation extent. As already adjudged, we use two of our better-performing modules pLSTM+softmax and pLSTM+sigmoid to represent the performance comparison with similar tasks for sarcasm, satire, and/or irony detection.

---
[1] https://www.gutenberg.org/

**Table 4.** Detailed information of the datasets, where *V* represents the lexical vocabulary length, *Size* is the cumulation of training and test size in MB of the respective data, and *SoA* is the respective state-of-art results observed. *SARC* represents *(tr + ts)*, i.e., training and test data combined. Dashed lines are introduced where no results are reported yet.

| Data | V | Size | Model | Train | Test | SoA |
|---|---|---|---|---|---|---|
| TCE[1] | 18020 | 100 | pLSTM+softmax | 98.90 | **98.31** | — |
|  |  |  | pLSTM sigmoid | 98.16 | **96.04** |  |
| TMB[1] | 69849 | 382 | pLSTM+softmax | 97.70 | 96.93 | — |
|  |  |  | pLSTM+sigmoid | 96.00 | 95.40 |  |
| SIGN [9] | 41097 | 2076 | pLSTM+softmax | 97.05 | 95.89 | — |
|  |  |  | pLSTM+sigmoid | 95.96 | 94.09 |  |
| Sar. V2 [10] | 43327 | 2568 | pLSTM+softmax | 95.37 | 94.06 | 76.00 [12] |
|  |  |  | pLSTM+sigmoid | 93.20 | 91.54 |  |
| SARC [11] | 51476 | 4646 | pLSTM+softmax | 94.91 | 93.00 | 77.00 [13] |
|  |  |  | pLSTM+sigmoid | 91.02 | 89.43 |  |

## 7 Discussion

Our proposed architecture can successfully detect sarcasm generation learned from the training epochs. All the LSTM modules in our architecture operate independently, drawing the same input sets. The performance of one such module does not affect the others. This provides room for monitoring the individual performances of each standalone module. Also, since modules are substructures of a cohesive architecture as a whole, the hyperparameters tuning and simulation environment is identical for all the modules. Finally, the statements are generated as a whole set combining all the modules' independent outputs. Besides maintain overall consistent performance, our method also misses out on certain occasions.

Also, we worked with an independent sarcasm corpus. Hence, we did not consider the context and situational sarcasm within regular conversations. The data is already rich with sarcastic dialogues and punchlines only. But sarcasm and irony within regular conversations are mostly reliant on the conversational context. As the training set of sarcastic statements are not part of any ongoing conversations, a few of them could actually sound generic ones and not sarcastic. Keeping it in mind, we tested our model on multiple other public sarcasm data as well, on which the previously applicable state-of-the-art works were also compared. It provided us a comparative concept of how our proposed approach fared on some of the standard sarcasm corpora available.

## 8 Conclusion and Future Work

Sarcasm can be produced through multi-parametric expressions. It is rather complex to understand at times; even for us humans. But when dealt with linguistics data only, we can exploit language-specific features in the likes of syntax, semantics, and vocabulary of that particular data. In this paper, we chose an open-sourced text corpus manually built on collecting some of the popular comedy show dialogues. We introduced the parallel deep LSTM network architecture, keeping in mind how the homogeneity of one such network modules would come up against each other when they are fed with the same inputs, identical tuning, but with different classifiers in outermost layers. To our expectations, two of the standalone modules performed well on the training data, achieving a maximum overall accuracy of 98.95%. We also traditionally tested our framework on popular public sarcasm corpora, and two of our independent modules scored over 95% accuracy in detecting sarcasm, with 98.31% being the highest accuracy obtained. However, for future endeavors, instead of deploying four deep neural models for achieving a homogenous task of sarcasm detection, we intend to develop a mimic the human alike statement succession initiating from random user input seed words to produce auto-generated natural sarcastic dialogues. For sentence-level sarcasm, there are a few areas of impact, which can be identified with the help of attention mechanisms. Considering it, we would like to build a persuasive model of the contextual conversation containing humor, wit, and irony for creating a sarcastic vocabulary entirely produced by deep neural networks.